%% file: acl.tex
\pdfoutput=1

\documentclass[11pt]{article}

\usepackage{acl}

\usepackage{times}
\usepackage{latexsym}

\usepackage[T1]{fontenc}

\usepackage[utf8]{inputenc}

\usepackage{microtype}
\usepackage{amsmath}
\usepackage{graphicx}
\usepackage{booktabs}

\usepackage{amssymb, amsthm}
\usepackage{enumitem}

\title{The Change that Matters in Discourse Parsing: \\ Estimating the Impact of Domain Shift on Parser Error}

\author{Katherine Atwell$^{\dagger \bigstar}$, \  Anthony Sicilia$^{\ddagger \bigstar}$, \ Seong Jae Hwang$^{\S}$, \ Malihe Alikhani$^{\dagger \ddagger}$ \\
 $^\dagger$Department of Computer Science and $^\ddagger$Intelligent Systems Program,
 University of Pittsburgh \\
 $^\S$Department of Artificial Intelligence, Yonsei University \\
 \texttt{\{kaa139, anthonysicilia\}@pitt.edu}, \\ \texttt{seongjae@yonsei.ac.kr}, \  \texttt{malihe@pitt.edu}}

\newtheorem{example}{Example} 
\newtheorem{theorem}{Theorem}

\newcommand{\customfootnotetext}[2]{{
  \renewcommand{\thefootnote}{#1}
  \footnotetext[0]{#2}}}

\begin{document}
\maketitle
\customfootnotetext{$\bigstar$}{K. Atwell and A. Sicilia contributed equally.}
\customfootnotetext{$\S$}{Work done while at University of Pittsburgh.}
\input{00-abstract}

\input{01-introduction} 
\input{02-related}
\input{03-data} 
\input{04-methods} 
\input{05-results} 
\input{06-conclusion} 

\section{Ethics}
Our experiments do not have any significant ethical concerns, as we do not work with any sensitive or personal data, nor do we work with human subjects; the datasets we use for our experiments are the PDTB 2.0 and 3.0, the RST Discourse Treebank, the GUM corpus, the TED-MDB, and the BioDRB.  Our work depends on pretrained models such as word embeddings.  These models are known to reproduce and even magnify societal bias present in training data.

\section*{Acknowledgements}
We would like to thank Amir Zeldes for his helpful feedback. 
Thanks to Pitt Cyber and DARPA grant prime OTA No. HR00112290024 (subcontract No. AWD00005100 and SRA00002145)
for partly supporting this
project. We also acknowledge the The
Center for Research Computing at the University
of Pittsburgh for providing the computational resources for many of the results within this paper. Any opinions, findings and conclusions or recommendations expressed in this material are those of the author(s) and do not necessarily reflect the position or policy of the U.S. Air Force Research Lab, DARPA, DoD and SRI International and no official endorsement should be inferred.

\clearpage

\bibliography{anthology,custom}
\bibliographystyle{acl_natbib}

\clearpage

\appendix
\input{07-appendix_a}
\input{08-appendix_b}
\input{09-appendix_c}
\clearpage
\input{10-appendix_d}
\clearpage
\input{11-appendix_e}
\input{12-appendix_f}
\input{13-appendix_g}
\input{14-appendix_h}

\end{document}

%% file: 00-abstract.tex
\begin{abstract}
Discourse analysis allows us to attain inferences of a text document that extend beyond the sentence-level. 
The current performance of discourse models is very low on texts outside of the training distribution's coverage, diminishing the practical utility of existing models. There is need for a measure that can inform us to what extent our model generalizes from the training to the test sample when these samples may be drawn from distinct distributions.
While this can be estimated via distribution shift, we argue that this 
does not directly correlate with change in the observed error of a classifier (i.e. error-gap).
Thus, we propose to use a statistic from the theoretical domain adaptation literature which \textit{can} be directly 
tied to error-gap. We study the bias of this statistic as an estimator of error-gap both theoretically and through a large-scale empirical study 
of over 2400 experiments
on 6 discourse datasets from domains including, but not limited to: news, biomedical texts, TED talks, Reddit posts, and fiction. 
Our results not only motivate our proposal and help us to understand its limitations, but also provide insight on the properties of discourse models and datasets which improve performance in domain adaptation. 
For instance, we find that non-news datasets are slightly easier to transfer to than news datasets 
when the training and test sets are very different. 
Our code and an associated Python package are available to allow practitioners to make more informed model and dataset choices.\footnote{\href{https://github.com/anthonysicilia/change-that-matters-ACL2022}{https://github.com/anthonysicilia/change-that-matters-ACL2022}}
\end{abstract}

%% file: 01-introduction.tex
\section{Introduction}
\label{sec:intro}
\begin{figure}[t]
    \centering
    \includegraphics[width=0.8\columnwidth]{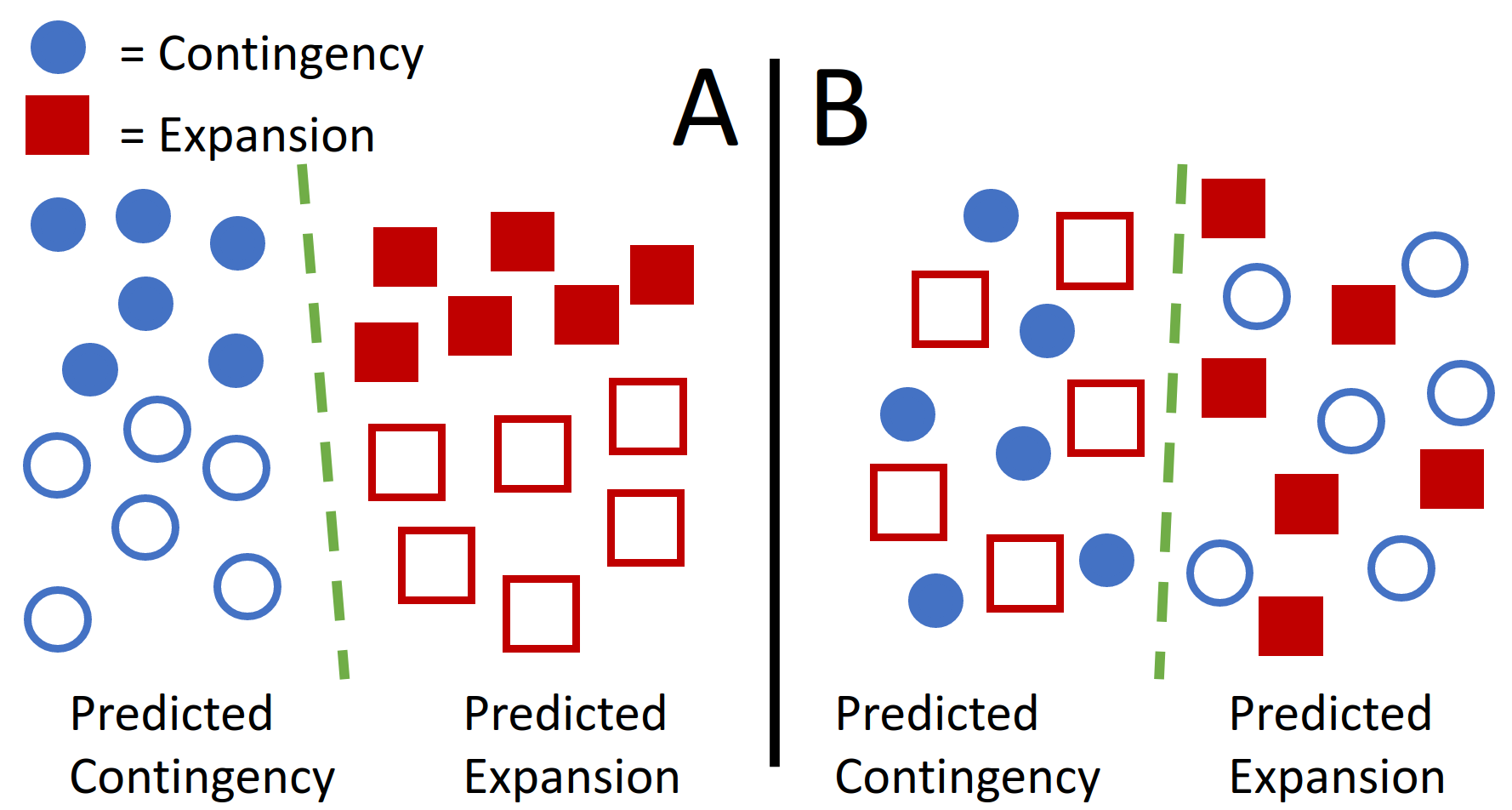}
    \caption{
    Solid/hollow shapes indicate training/test set, while circles/squares indicate the correct labels.
    \textsf{(A)} Vertical shift is easily identified, but the classifier (dotted line) does well on both domains. \textsf{(B)} In the feature space, shift is imperceptible, but the classifier assigns the incorrect relation label to each point in the test set. In both, identifiable shift does \textit{not} correlate with 
    the classifier's ability to correctly predict the discourse relation}
    \label{fig:intro}
    \vspace{-0.5em}
\end{figure}
Coherence analysis of text is a key area of natural language processing.
Discourse parsing models are trained on a dataset annotated according to a \textit{discourse framework}, wherein the discourse structure and how the discourse units are connected are identified and labeled. Some discourse frameworks \cite{PDTB2004, PDTB2008, webber2019penn} focus on shallow relations between two individual discourse units, while others \cite{carlson2003building, lascarides2008segmented} focus on learning a more hierarchical structure. Discourse models have been shown to improve performance in several fundamental NLP tasks, such as summarization \cite{marcu1999discourse, marcu2000theory, cohan-etal-2018-discourse}, sentiment analysis \cite{bhatia-etal-2015-better}, machine comprehension \cite{narasimhan2015machine}, and machine translation \cite{guzman2014using}. However, in some cases, using discourse relations themselves has been found not to improve, or even to hurt, performance in other tasks when learning the coherence structure of text seems critical\cite{zhong2020discourse, feng2015rst}. There are several possible reasons for this: due to the difficulty of the annotation task, datasets labeled with these discourse relations are typically small, and the most widely used datasets consist only of news texts. As a result, the performance of discourse models trained on these datasets is very low, and even slight domain shift has been shown to worsen the performance \cite{atwell2021we}. 
Thus, for the task of discourse parsing, it is especially important to be cognizant of the effects of domain shift, and choose models and training datasets that are likely to generalize well on the target domain. 

To estimate the extent of a model's generalizability on a particular train/test pair, common proposals suggest using two-sample statistics which capture distributional shift in the feature space \citep{rabanser2019failing}.
However, the working hypothesis of this paper is that \textit{changes in feature-distribution do not necessarily equate to changes in a classifier's error}; i.e., from train to test sample. Figure~\ref{fig:intro} captures this idea by illustrating some examples in simple 2D-space where domain shift may occur without high error, and vice versa, in the context of discourse parsing.

Motivated by this hypothesis, we look to existing theoretical domain adaptation literature. We propose to use a statistic which has not only been designed to incorporate information about the classifier we would like to transfer, but has also been shown (theoretically) to directly relate to model performance on the test set. Namely, we consider generalization of the source-guided discrepancy \citep{kuroki2019unsupervised} which we call the $h$-discrepancy defined for any classifier $h$ (we introduce and define this metric in Section~\ref{sec:methods}). We provide novel theoretical analysis of the errors of this statistic in estimating adaptation performance and, based on this, hypothesize this statistic will correlate more substantially with the classifiers' generalization ability than the two-sample statistics previously mentioned. We support this hypothesis by illustrating these correlations across several different widely-used discourse datasets (described in Section \ref{sec:experiments}). We also provide a detailed empirical analysis of the estimation error of this statistic in predicting adaptation performance using a regression model. 
In doing so, we provide insights on the effect of various properties of different discourse models and datasets on performance in domain adaptation, which we enumerate in Section \ref{sec:conclusion}. We expand on these contributions next.

First, we contribute a new theoretical analysis to characterize the bias of the $h$-discrepancy as an estimator of performance in domain adaptation. Although this discrepancy is typically biased, we provide upper and lower bounds on this bias and interpret them to provide insight on the use of this statistic in practice. In particular, we show that a small $h$-discrepancy often means the practitioner can be confident in transferring the model from the train- to the test-set. Our theoretical analysis motivates our hypothesis that the $h$-discrepancy 
should outperform common two-sample statistics.

Next, we empirically study the aforementioned hypothesis. We compare correlation of the $h$-discrepancy with performance in domain adaptation against correlation of various two-sample statistics across multiple discourse datasets. As we are aware, this large-scale comparison has never been done for discourse relation classification. As mentioned above, the results of this analysis provide support for our hypothesis that the $h$-discrepancy is the best estimator of performance changes under domain shift. As such, we argue that computational discourse practitioners should use this statistic to determine the model/dataset likely to maximize performance under domain shift.

We also perform a regression analysis of the estimation errors of the $h$-discrepancy as an estimator for domain adaptation performance. This analysis allows us to understand the properties and pitfalls of our estimator. Further, it allows us to gain useful insights into how different types of datasets, genres, feature representations, and models influence the generalizability of discourse parsers. We enumerate these insights and discuss their implications for discourse researchers in Section \ref{sec:results}.  


In the sections below, we further discuss and motivate the need for domain-adaptation bounds tied directly to the error gap for more informed insights into performance gaps under domain shift. We hope that discourse researchers use our results, and our code, as a starting point for model and dataset selection in their own studies.

%% file: 02-related.tex
\section{Related Work}
\label{sec:related}

\subsection{Discourse and Domain Shift}
Computational analysis of discourse has been the focus of several shared tasks \cite{xue2015conll, xue2016conll, zeldes-etal-2019-disrpt, disrpt-2021-shared}, and there have been several discourse-annotated corpora for multiple languages \cite{zeyrek2008discourse,meyer2011multilingual,danlos2012vers,zhou2015chinese,zeyrek2020ted, da-cunha-etal-2011-development, das2018developing, afantenos2012empirical}. Despite their widespread use, implicit sense classification remains a challenging task \cite{liang-etal-2020-extending}, and discourse models have been shown not to perform well under even gradual domain shift \cite{atwell2021we}, which may be the result of the limited timeframe and distribution of the articles contained in the most commonly used English discourse datasets, the Penn Discourse Treebank \cite{PDTB2004, PDTB2008, webber2019penn} and the RST Discourse Treebank (RST-DT) \cite{carlson2003building}. These datasets are both made up of Wall Street Journal articles spanning a three-year period, and thus do not contain much variation with respect to linguistic distribution. 

Several works have quantified domain shift in the context of natural language processing, mostly in the task of sentiment analysis. For instance, \citet{plank2011effective} use word frequencies and topic models to measure domain similarity, while \citet{wu2016sentiment} use sentiment graphs. In contrast, ours is the first to consider quantifying domain shift in discourse analysis.
With respect to our methodology, some works take a similar approach.
\citet{blitzer2007biographies} and \citet{elsahar-galle-2019-annotate} also use a statistic from domain adaptation theory, employing the $\mathcal{H}$-divergence to analyze a sentiment classification task on the Amazon Reviews dataset, while \citet{ruder2017data} use $\mathcal{H}$-divergence to select the source datasets for transfer. However, none of these works have studied the $h$-discrepancy we study here, which is dependent on the classifier used for inference. In comparison, the $\mathcal{H}$-divergence ignores information about the model we would like to transfer, and therefore, will be less sensitive (e.g., in model-selection contexts).

To the best of our knowledge, no works have yet studied the correlation of statistics from the theoretical domain adaptation literature with the adaptation performance of discourse parsers. This is especially true given the wide array of different datasets and distributional shifts we consider as well as the theoretical and empirical tools we propose to conduct our study. Both our novel theoretical result (Theorem~\ref{thm:main}) and our large-scale regression analysis (Section~\ref{sec:results}), provide new, practical insights on domain-shift in discourse parsing.

\subsection{Domain Adaptation Theory}
Statistics that relate to domain adaptation performance have long been studied in the theoretical literature. \citet{kifer2004detecting, ben2007analysis, ben2010theory} initiate this investigation with a modification of the total variation distance (the $\mathcal{H}$-divergence) that depends on the set of classifiers $\mathcal{H}$; this statistic can be directly related to adaptation performance through a finite sample bound. \citet{mansour2009domain} extend this discussion from classification error to general loss functions. Certain two-sample statistics can also be related to adaptation performance through finite sample bounds, but only under stringent assumptions on the space of classifiers and the computation of the two-sample statistic \citep{NIPS2009_685ac8ca, gretton2012kernel, long2015learning, Redko2020ASO}. 

Assumptions, in general, play a large role in successful domain adaptation. In fact, common adaptation algorithms can actually \textit{worsen} performance if important assumptions are not met \cite{zhao2019learning, wu2019domain}. Different assumptions have led to diverse theories disjoint from the $\mathcal{H}$-divergence, including proposals of \citet{lipton2018detecting}, \citet{johansson2019support}, and \citet{combes_and_zhao}. Under certain strict and untestable assumptions, it is even possible to derive unbiased estimators of adaptation performance \citep{sugiyama2007covariate, you2019towards}. We later discuss our own assumptions on the \textit{adaptability} $\lambda$ which are typical when using the $\mathcal{H}$-divergence and its descendants. We find these assumptions to be comparatively mild. In comparison to some others, they have also been theoretically argued to be of vital importance \citep{david10a}.

%% file: 03-data.tex
\section{Methods}
\label{sec:experiments}



\paragraph{Data}
Our English datasets are all based on either the RST Discourse Treebank or Penn Discourse Treebank frameworks, which we describe in Appendix \ref{sec:frameworks}. Table \ref{tab:discourse_datasets} summarizes differences between the datasets we use in our experiments. 

\begin{table}
    \centering
    \small
    \begin{tabular}{p{4cm}|p{1cm}p{1cm}}
       Dataset  & Genre & Label schema  \\
         \toprule
        RST-DT \cite{carlson2003building} & News & RST-DT \\
        PDTB 2.0 \cite{PDTB2008} & News & PDTB \\
        PDTB 3.0 \cite{webber2019penn} & News & PDTB \\
        BioDRB \cite{ramesh2010identifying} & Bio & PDTB \\
        TED-MDB \cite{zeyrek2020ted} & TED talks & PDTB \\
        GUM \cite{Zeldes2017} & Multiple & RST-DT 
    \end{tabular}
    \caption{Characteristics of each discourse dataset used in our study. The "multiple" domains in the GUM corpus are as follows: Academic, Biography, Fiction, Interview, News, Reddit, Travel, and How-to guides. The main distinction between the PDTB-2 and PDTB-3 is the presence of intra-sentential implicit discourse relations in the PDTB-3.}
    \label{tab:discourse_datasets}
\end{table}



\paragraph{Features}
For each discourse relation, we encode the argument pair as features. For the RST-DT and GUM corpus, we thus only use discourse relations between two EDUs. To encode argument pairs, we concatenate and tokenize them using the BERT \cite{devlin-etal-2019-bert} tokenizer. We then feed these tokens through the pretrained base BERT model and experiment with two different ways of capturing the model output: using the pooled output, e.g. the output of the [CLS] token, and averaging the hidden states. We will refer to these encodings as P-BERT and A-BERT respectively. We also experiment with encoding our argument pairs using SentenceBERT \cite{reimers2019sentence} which we will refer to as S-BERT. 

\paragraph{Label Set}
For the datasets with the PDTB label schema, we use only the top-level sense labels (Expansion, Contingency, Comparison, and Temporal). We use the top-level RST-DT classes for the datasets with the RST-DT label schema, and map the GUM corpus classes to the RST-DT classes using \citet{braud2017cross}.
We recognize this mapping will not be perfect, as mappings between frameworks rarely are, but we follow the mapping with empirical support from \citet{demberg2017compatible} and focus on the predicting top-level relations between two discourse units. As a consequence, we expect to observe distinct labeling functions (i.e., annotator decisions) across domains from separate discourse frameworks. 

\paragraph{Experiments}
Each data point in all of our results (e.g., when computing correlation or doing regression analysis) corresponds to a particular \textit{experiment} done on a source (train) dataset $S$ and target (test) dataset $T$ using a classifier $h$. The classifier $h$ is trained on the source $S$ and evaluated on target $T$. This is meant to mimic a common domain adaptation scenario in which the NLP practitioner would like to transfer a pre-trained discourse classification model to a new unlabeled dataset (i.e., this is discussed again in Section \ref{sec:methods}). For each experiment, $h$ is trained using a standard optimization procedure to have low error on $S$. We discuss this procedure and its competitiveness with respect to the state-of-the-art in Section~\ref{sec:results}. 

For each dataset, we randomly split the dataset in half based on 3 different seeds. For example, PDTB 2.0 (10K examples) is randomly split into to disjoint sets of about 5K examples. 
The pair $S$ and $T$ are taken from the set of these splits using each of the different BERT representations. We restrict the pair to have a common set of discourse labels. For example, we only transfer from $S$ using the PDTB label schema to $T$ using the same schema. 

For experiments involving PDTB label schema, we consider single-source domain adaptation, which simply pairs one data split $S$ with another $T$. For instance, the first half of the TED-MDB and the second half of the BioDRB, or, the first half of BioDRB and the second half of BioDRB. 

For experiments involving RST-DT label schema, we use both single-source and multi-source domain adaptation setups. We use the multi-source setup for domains in the GUM corpus. Here, $T$ is derived from a single domain and $S$ from all of the other domains contained in the corpus (i.e., $S$ would contain 7 of the GUM domains and $T$ would contain the remaining one). Although we continue to split the domains in half, we only use one of the halves in this case to prevent samples from the target distribution from appearing in the source. We use the single-source setup for RST itself. Here, $S$ is one split of RST while $T$ is another.

Importantly, experimenting with this variety of setups allows us to simulate variability arising from sampling as well as study different degrees of domain shift. Accounting for each pair and each random seed for model training, the number of $(S,T,h)$ triples we study totals more than 2400.




%% file: 04-methods.tex
\section{Quantifying Meaningful Domain Shift}
\label{sec:methods}
Identifying and quantifying domain shift is a classical problem. Perhaps, the most widely used mechanism for this task is the two-sample test; i.e., a test designed to indicate difference of distribution between two samples. We begin this section by discussing a few of the statistics used in these tests. We observe a common problem in using these statistics to predict adaptation performance, and following this, discuss the aforementioned $h$-discrepancy. 

\subsection{Common Two-Sample Test Statistics}
We now informally discuss some common statistics used in two-sample tests. These statistics can be easily adapted to infer adaptation performance under the assumption that changes in distribution perfectly correlate with changes in error. As mentioned earlier, we do \textit{not} agree with this hypothesis. Still, these types of statistics serve as a good point of comparison. In our experiments, we compute each of these statistics using the PyTorch library \texttt{torch\_two\_sample} \citep{torchtwosample}. 
{\setlist{nolistsep}
\begin{itemize}[noitemsep]
    \item \textbf{FRS}: \citep{friedman1979multivariate} counts edges from $S$ to $T$ in a graph representation.
    \item \textbf{Energy}: \citep{szekely2013energy} compares dissimilarity of points within/across $S$ and $T$.
    \item \textbf{MMD}: \citep{gretton2012kernel} compares \textit{similarity} of points within/across $S$ and $T$.
    \item \textbf{BBSD}: \citep{lipton2018detecting} applied MMD to softmax output (i.e., scores) of classifier $h$.
\end{itemize}}
For more computational details, see Appendix~\ref{sec:two-sample-details}. 

\paragraph{A Common Problem}
The majority of these statistics share the common trait that they were originally designed to test differences in feature distribution -- \textit{not} differences in hypothesis error. As such, while we do expect them to be sensitive to changes in error -- in so far as changes in feature distribution relate to changes in error -- we have no theoretical reason to expect this should be the case. As we saw in Figure~\ref{fig:intro}, these two changes can be very different: large changes to the distribution of features may not hurt performance in every case and imperceptible changes to the distribution of features can have large impact when the labeling function changes. In fact, most of these statistics do not even incorporate information about the classifier we use for inference. While BBSD does, we are not aware of any theoretical arguments linking it to adaptation performance in the same way as the $h$-discrepancy (discussed next).

\subsection{Identifying the Change that Matters}
Contrary to those statistics described above, the statistic we give in this section is directly related to adaptation performance by theoretical means. Before beginning our description of this metric, we need to formalize our mathematical setup and a particular notion of adaptation performance.

\paragraph{Mathematical Setup}
We measure adaptation performance through the error-gap which is defined:
\begin{equation}\small
   \Delta_h(S, \mathbb{T}) = \left |\mathbf{R}_S(h) - \mathbf{R}_{\mathbb{T}}(h) \right |
\end{equation}
where $S$ is a sample and $\mathbb{T}$ is a distribution -- both over a space $\mathcal{X} \times \mathcal{Y}$. In this paper, $\mathcal{X}$ is usually the space of real-valued vectors (i.e., BERT representations for argument pairs) and $\mathcal{Y}$ corresponds to a set of possible discourse labels. $h$ is a classifier $h : \mathcal{X} \to \mathcal{Y}$ and the risk $\mathbf{R}_\mathbb{D}(h)$ is defined for distribution $\mathbb{T}$ as $\mathbf{R}_\mathbb{T}(h) = \mathbf{Pr}(h(\tilde{X}) \neq \tilde{Y}), \ (\tilde{X},\tilde{Y}) \sim \mathbb{T}$.
For sample $S = (X_i, Y_i)_{i=1}^n$, we instead write $\mathbf{R}_S(h) = n^{-1} \sum\nolimits_i 1[h(X_i) \neq Y_i]$
where $1[\cdot]$ is the indicator function. To compute each statistic which we would like to use to infer the error-gap, we assume access to the mentioned sample $S$ drawn i.i.d from some distribution $\mathbb{S}$. We also assume access to a new unlabeled sample $T_{X} = (\tilde{X}_i)_{i=1}^m$ drawn i.i.d from the $\mathcal{X}$-marginal $\mathbb{T}_{X}$ of the distribution $\mathbb{T}$. In general, we do not know whether $\mathbb{T} \neq \mathbb{S}$ or $\mathbb{T} = \mathbb{S}$, but may have reason to suspect $\mathbb{T} \neq \mathbb{S}$. 

\paragraph{Roadmap} In the next part, we give the statistic we would like to use to predict adaptation performance. We then quantify its bias as an estimator for the error-gap with a theoretical result. We also propose a technique to study the relationship between this statistic and the error-gap empirically through a regression analysis. Finally, we show how this technique can be used to study the impact certain attributes of a model or dataset have on error-gap. 

\paragraph{Source-Guided Discrepancy}
The source-guided discrepancy was proposed by \citet{kuroki2019unsupervised} with a similar conceptualization given independently by \citet{zhang2019bridging}. These statistics improve upon a long history of domain adaptation statistics \citep{kifer2004detecting, blitzer2007biographies, ben2007analysis, ben2010theory}, specifically, by incorporating information on the source-labels.
We consider a generalization of the source-guided discrepancy which we call the $h$-discrepancy, defined for any classifier $h$. For samples $S$ and $T_X$, a binary label space $\mathcal{Y}$, a space of classifiers $\mathcal{H}$ over $\mathcal{X} \times \mathcal{Y}$, and any\footnote{The source-guided discrepancy originally proposed by \citet{kuroki2019unsupervised} considers only one particular $h$.} fixed classifier $h \in \mathcal{H}$, it is defined as:
\begin{equation}\small
\begin{split}\label{eqn:D}
D & = \max\nolimits_{g \in \mathcal{H}} \lvert \mathbf{R}_U(g) - \mathbf{R}_V(g) \rvert \quad\text{where} \\
U & = ((X_i, h(X_i))_{i=1}^n, \quad V = ((\tilde{X}_i, h(\tilde{X}_i))_{i=1}^m,
\end{split}
\end{equation}
and recall, $S_X = (X_i)_i$ and $T_X = (\tilde{X}_i)_i$. In the binary case, \citet{kuroki2019unsupervised} show that this may be approximated by learning a classifier (i.e., $g$) which \textit{agrees} with $h$ on the source sample $S_X$ and \textit{disagrees} with $h$ on the target sample $T_X$. Their procedure extends naturally to the multi-class case as well, but we must disambiguate between the possible ways in which $g$ can disagree with $h$. In our experiments, we do so by training $g$ to pick the next most likely label according to the scores of $h$. For a better approximation, one should compute $D$ again, reversing the roles of $S / T$ and taking the larger of the values as the final result. With binary labels, the two values will often coincide, but this should not be assumed in multi-class settings. 
\paragraph{Theoretical Motivation}
Here, we provide our primary motivation for the $h$-discrepancy as an estimator of error-gap. Our result makes use of the work of \citet{crammer2007learning}, \citet{ben2010theory}, and \citet{kuroki2019unsupervised}. It distinguishes itself from these finite sample bounds in that it explicitly concerns itself with the bias of $D$ as an estimator of error-gap. Proof is given in Appendix~\ref{sec:proof}.
\begin{theorem}\label{thm:main}
Let $\mathcal{Y}$ be a binary space and let $\mathcal{H}$ be a subset of classifiers in $\mathcal{Y}^\mathcal{X}$. Then, for any realization of $S$, for all $h \in \mathcal{H}$,
\begin{equation}\small
 -\mathbf{E}_T[\lambda] \leq \mathbf{E}_T[D] - \Delta_h(S,\mathbb{T}) \leq \mathbf{E}_T[D]
\end{equation}
where $\lambda = \min_{h' \in \mathcal{H}} \mathbf{R}_S(h') + \mathbf{R}_T(h')$. 
\end{theorem}
Notice, when $\mathbf{E}[\lambda]$ is small and $\mathbf{E}[D]$ is also small we know the bias must be small because it is ``sandwiched'' between these two.
In this situation, the practitioner can very confidently transfer $h$ from $S$ to $T$. In practice we cannot compute $\lambda$ since it requires labels from $T$, still we often expect $\mathbf{E}[\lambda]$ to be small. In particular, this term is often called the \textit{adaptability} as it captures irreconcilable differences between the source and target labeling functions. In discourse, such differences are primarily determined by the discourse framework and annotator. As first observed by \citet{ben2010theory} (i.e., concerning a similar term), $\lambda$ is small whenever there is \textit{any} classifier in $\mathcal{H}$ which does well on $S$ and $T$ simultaneously. If $S$ and $T$ come from the same discourse framework, this should not be difficult for sufficiently complex $\mathcal{H}$. Even if $S$ and $T$ come from distinct discourse frameworks, this is still not an overly strong requirement because neural-networks, for example, have been shown to perfectly fit even random labeling \citep{zhang2016understanding}. Thus, in many cases,\footnote{One should be cautious of broad generalizations in adaptation, since failure to carefully consider $\lambda$ can be disastrous for algorithm design \citep{zhao2019learning, johansson2019support}.} we are primarily concerned with the positive bias of $D$. When $\mathbf{E}[D]$ is larger, the positive bias of $D$ can also be larger. Intuitively, $D$ might have more ``false positives'' where it reports a high value but the error-gap is actually comparatively small. In this sense, it is a conservative statistic. It plays things on the ``safe side.'' So, while $D$ will possibly have some bias, it is at least described by the above bounds. As we are aware, the two-sample statistics discussed previously do not have such a description. 

\paragraph{Regression Analysis of Errors of $D$}
From Theorem~\ref{thm:main}, we do not expect the random estimation error $D - \Delta_h(S, \mathbb{T})$ to be zero. So, in our experimentation, we propose to study this quantity through a regression analysis. Namely, suppose $\mathbf{X} \in \mathbb{R}^{N \times p}$ is some fixed, non-singular design matrix whose rows each represent one of $N$ experiments and whose columns represent one of $p$ features for each experiment. An experiment corresponds to an $(S,T,h)$ triple as disucssed in Section~\ref{sec:experiments}. The features are dependent on properties of the datasets and models used in each experiment as well as realizations of $h$-discrepancy, adaptability, and training error. Then, we assume
\begin{equation}\label{eqn:reg}\small
    \mathbf{Y} = \mathbf{X}\boldsymbol{\beta} + \boldsymbol{\epsilon}
\end{equation}
where the randomness in the outcome $\mathbf{Y}$ comes from $\boldsymbol{\epsilon}_i \overset{\mathrm{i.i.d.}}{\sim} N(0, \sigma^2)$, $\sigma > 0$. The response $\mathbf{Y} = (D_i - \Delta_h(S, \mathbb{T})_i)_{i=1}^N$ are realizations of estimation error across $N$ experiments.\footnote{We do not have access to $\mathbb{T}$, so we use sample $T$ instead.} We give model diagnostics and details of the design matrix $\mathbf{X}$ in Appendix~\ref{sec:reg-diag}; it is selected manually using domain knowledge and to meet model assumptions. 

Regression analysis is particularly useful because standard techniques allow us to understand and isolate the impact of individual columns (i.e., features) in $\mathbf{X}$ on the estimation errors of $D$. In particular, we can use this model to determine the expected change in estimation error as a function of a particular feature, while controlling (i.e., holding constant) all other features in $\mathbf{X}$:
\begin{equation}\small \label{eqn:small-reg-eg}
    \mathbf{E}[\mathbf{Y}_i \mid \mathbf{X}_i = \mathbf{x}] - \mathbf{E}[\mathbf{Y}_i \mid \mathbf{X}_i = \mathbf{x}']
\end{equation}
where $\mathbf{x}$ is any setting of the features and $\mathbf{x}'$ is identical to $\mathbf{x}$ except every component involving the feature of interest is modified (e.g., increased) systematically. For a specific example using Eq.~\eqref{eqn:small-reg-eg}, consider inspecting the change in estimation error as a function of increase in $h$-discrepancy (controlling for all other features). In this case, Eq.~\eqref{eqn:small-reg-eg} evaluates to a polynomial\footnote{For details, please see Appendix~\ref{sec:reg-eg}, Example~\ref{examp:reg-1}.} in the coefficients $\boldsymbol\beta$ and components of $\mathbf{x}'$, so we can estimate this result in an unbiased manner using the OLS estimate $\boldsymbol{\hat{\beta}} = (\mathbf{X}^\mathrm{T}\mathbf{X})^{-1}\mathbf{X}^\mathrm{T}\mathbf{Y}$. To empirically validate our theoretical analysis, we might check if this polynomial is an increasing, positive function; i.e., because our theory predicts increases in the expected $h$-discrepancy allow for increases in bias.

\paragraph{Regression Analysis of Error-Gap}
Given $\mathbf{X}$ and $\boldsymbol\beta$, rearranging Eq.~\eqref{eqn:reg} lets us also write
\begin{equation}\small
\label{eqn:reg-2}
    \Delta_h(S, \mathbb{T})_i = D_i - \mathbf{X}_i\boldsymbol{\beta} + \boldsymbol{\epsilon}_i
\end{equation}
where $\mathbf{X}_i$ is the $i^\text{th}$ row of $\mathbf{X}$; i.e., the features of the $i^\text{th}$ experiment. Similar to before, this type of analysis lets us draw interesting insights. In particular, we can isolate the impact of features in $\mathbf{X}$ on the error-gap. Since our design matrix $\mathbf{X}$ controls for training error, the error-gap can be interpreted to act as a measure of performance in domain adaptation (DA). Those features which are positively associated with error-gap can be said to be worse for DA. Likewise, those with negative association are ``better'' for DA. As before, we isolate the impact of a feature by checking the change in error-gap as a function of change in this feature (i.e., similar to Eq.~\ref{eqn:small-reg-eg}). 
Appendix~\ref{sec:reg-eg} Example~\ref{examp:reg-2} uses this technique to isolate the impact of different BERT representations on error-gap. 

%% file: 05-results.tex
\section{Results}
\label{sec:results}
\begin{table*}[t!]
\centering
\small
\begin{tabular}{ c|ccccc|ccccc}
\toprule
&\multicolumn{5}{c|}{\textbf{Spearman (Rank) Correlation}} 
&\multicolumn{5}{c}{\textbf{Pearson (Linear) Correlation}} \\
\textbf{Split} & \textbf{FRS} & \textbf{Energy} & \textbf{MMD} & \textbf{BBSD} & $h$-\textbf{disc} & \textbf{FRS} & \textbf{Energy} & \textbf{MMD} & \textbf{BBSD} & $h$-\textbf{disc} \\
\toprule
All & 0.5394 & 0.6059 & 0.5051 & 0.4054 & \textbf{0.8299} & 0.4986 & 0.4396 & 0.3413 & 0.4004 & \textbf{0.7628} \\
\midrule
PDTB & 0.5451 & 0.6359 & 0.5472 & 0.4746 & \textbf{0.8265} & 0.5295 & 0.4704 & 0.3709 & 0.4274 & \textbf{0.7642} \\
RST-DT & 0.2166 & 0.3059 & -0.0011 & 0.2087 & \textbf{0.7625} & 0.2853 & 0.1660 & -0.1605 & 0.1677 & \textbf{0.7599} \\
\midrule
News & 0.5262 & 0.6356 & 0.5507 & 0.5759 & \textbf{0.8517} & 0.7079 & 0.6302 & 0.5558 & 0.5386 & \textbf{0.8890} \\
Other & 0.3760 & 0.4517 & 0.2767 & 0.1737 & \textbf{0.8386} & 0.3420 & 0.2791 & 0.1760 & 0.2051 & \textbf{0.7072} \\
\midrule
WD & 0.0884 & 0.5735 & -0.0324 & 0.2368 & \textbf{0.7890} & 0.1075 & 0.5831 & -0.0515 & 0.4853 & \textbf{0.9519} \\
OOD & 0.4597 & 0.5249 & 0.3917 & 0.2813 & \textbf{0.7666} & 0.4342 & 0.3909 & 0.2761 & 0.3745 & \textbf{0.6976} \\
\bottomrule
\end{tabular}
\caption{Correlations with error-gap for each statistic. Data splits indicate the subset of data used. 
$h$-discrepancy consistently yields the largest correlation with error-gap; i.e., difference in Pearson correlations are all significant at level $\alpha = 0.001$ using test of \citet{steiger1980tests} implemented by \citet{diedenhofen2015cocor}.}
\label{table:correl}
\end{table*}

\begin{figure*}
    \centering
    \includegraphics[width=\textwidth]{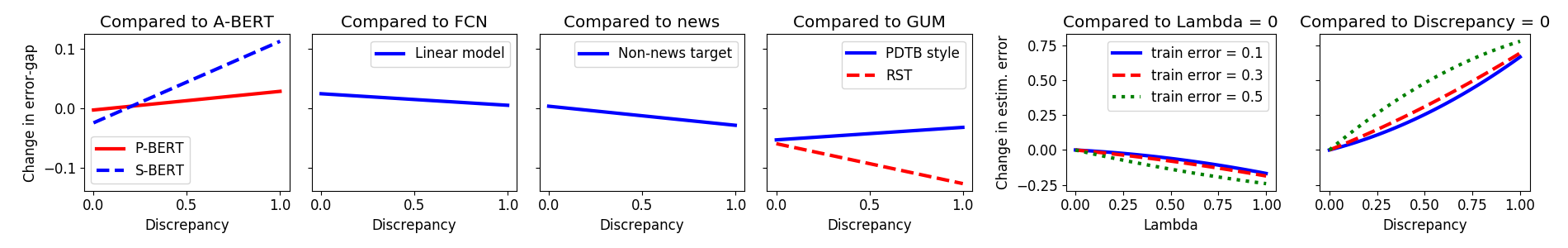}
    \caption{(Left, 1-4) Expected change in error-gap when changing properties of the dataset or model. Shown as a function of discrepancy and controls for all other features of the experiment.  Reference category is indicated in title. (Right, 5-6) Expected change in estimation error of $h$-discrepancy shown as a function of $\lambda$ (5th) and discrepancy (6th). Left assumes use of A-BERT and FCN on a GUM non-news target, but trends are consistent in other cases.}
    \label{fig:all}
\end{figure*}

\subsection{Analysis of Transfer Error}

\paragraph{Comparison to Other Work}
\emph{Our experimental setup produces results comparable to current discourse models.} In Appendix \ref{sec:model-training}, Figure \ref{fig:transfer_error_wd_ood} shows the distribution of the error rates when transferring on within- and out-of-distribution datasets. To validate whether our setup is comparable to other discourse parsing models, we compare error rates to current implicit sense classifiers; e.g., \citet{kishimoto-etal-2020-adapting} who achieve an error rate of $\approx 0.38$ under a comparable setup. Our PDTB within-distribution results often improve upon this.

\paragraph{Error Analysis across Genres}
\emph{Fiction and How-To Guides are the most difficult to transfer to, while Academic Journals and Biographies are the easiest.} Figure \ref{fig:transfer_error_gum} in Appendix \ref{sec:model-training} shows the error rates for multi-source adaptation on the GUM corpus across S-BERT, P-BERT, and A-BERT. Although the error rates differ across these three representations, the relative order of the GUM corpus domains with respect to transfer error is fairly consistent across all of them. For all three, the highest mean error rate occurred in the How-to Guide and Fiction domains, and the lowest mean error rate occurred in the Academic and Biography domains. 

\subsection{Analysis of Correlations}
In Table~\ref{table:correl}, we show linear and rank correlation of each statistic with the error-gap. This tests the ability of each statistic to discern scenarios where domain adaptation performance may be either good or bad. In practice, a statistic with good rank correlation can be used in model-selection or (source) dataset selection. A statistic with good linear correlation may also be used and will be easier to interpret since we expect changes in the statistic to be proportional to changes in the error-gap.

\paragraph{Comparison of Statistics}
\emph{$h$-discrepancy is consistently, most strongly correlated with error-gap}. The overarching trend is that the $h$-discrepancy is far better than every other statistic with regards to both types of correlation. In fact, the linear correlations are not much worse than the rank correlations (in some cases they are even better). This validates our opening hypothesis that domain-shift does not always correlate with domain adaptation performance (i.e., error-gap). It is important to also consider the classifier we use. Still, BBSD -- another statistic that relies on the classifier -- is also somewhat ineffective compared to the $h$-discrepancy. Importantly, despite depending on the classifier, BBSD was still designed with identification of feature-distribution shift in mind. In some sense, this observation validates our theoretical motivations for the $h$-discrepancy (i.e., Theorem~\ref{thm:main}) which directly relates it to error-gap. Our results indicate that, at least for the task of discourse parsing, $h$-discrepancy is the most effective statistic to use with regards to predicting error-gap.

\paragraph{Additional Trends} \emph{Experiments using RST-DT label schemas and non-news targets show very low correlation between distributional shift and error-gap}.
If we look at particular experiment subsets, we also see some interesting trends. First, most statistics are better correlated with error-gap datasets that use the PDTB label schema than those that use the RST-DT label schema. The difference is less pronounced for the $h$-discrepancy than for the other statistics, suggesting that it is especially important to use statistics tied directly to the error-gap when working with datasets that use the RST-DT schema. The same is true when the test dataset is comprised of news articles instead of other types of text. 

\textit{The $h$-discrepancy has highest linear correlation on similar distributions.} We observe much stronger linear correlation between the $h$-discrepancy and error-gap on within-distribution adaptation scenarios (\textbf{WD}) as compared to out-of-distribution adaptation scenarios (\textbf{OOD}). We believe this is because the $h$-discrepancy is typically small when $S$ and $T$ follow a similar distribution. As Theorem~\ref{thm:main} notes, the bias of the $h$-discrepancy as an estimator for error-gap can be near zero if both $\mathbf{E}[D]$ and $\mathbf{E}[\lambda]$ are small; i.e., we expect the linear correlation of a nearly unbiased estimator to be fairly high. 
\subsection{Regression Analysis of Estimation Error}
Figure~\ref{fig:all} shows expected change in estimation error of $h$-discrepancy (used as an estimator for error-gap). Trend lines indicate expected change as a function of the adaptability $\lambda$ and the discrepancy $D$ compared to the case where each is 0.\footnote{Note, if both are 0 in expectation, $D$ is unbiased.} Trends are computed using a similar technique for regression analysis as described in Appendix~\ref{sec:reg-eg} Example~\ref{examp:reg-1}. The takeaway is that these empirical results are consistent with our theoretical discussion surrounding Theorem~\ref{thm:main}. As $\lambda$ increases, the estimation error decreases. Similarly, Theorem~\ref{thm:main} predicts the possibility of negative bias when $\lambda$ is large. As $D$ increases, the estimation error does the same. Theorem~\ref{thm:main} agrees here too, predicting the possibility of positive bias when $D$ is large.%
\subsection{Regression Analysis of Error-Gap}
Figure~\ref{fig:all} also shows expected change in error-gap when modifying categorical features of the experiment; e.g., use of S-BERT vs. A-BERT. Trend lines indicate expected change as a function of $h$-discrepancy and are computed using a similar technique for regression analysis as described in Appendix~\ref{sec:reg-eg} Example~\ref{examp:reg-2}. Since we control for training set error, positive changes in error-gap indicate a setting is better for domain adaptation, while negative indicates the opposite. This regression analysis also controls for changes in discourse framework using explicit indicator variables as well as the term $\lambda$ (see discussion after Theorem~\ref{thm:main}).

\paragraph{BERT features}
\emph{S-BERT is better for similar train and test sets, while A-BERT is better for more divergent sets}. As a function of discrepancy, S-BERT is better for DA when the discrepancy is small. As the difference between the train and test set increases, the reference category (i.e., A-BERT) is better for DA. Comparing P-BERT to A-BERT we do not see large differences; marginally, A-BERT is better as discrepancy increases. These results are consistent with typical rules of thumb on model complexity. A more complex feature representation (i.e., S-BERT or P-BERT) is beneficial when training and test distributions align, but allows for overfitting when discrepancy increases.

\paragraph{Classifier}
\emph{Linear classifiers perform marginally worse than neural-networks}. 
In general, fully-connected networks (FCNs) appear to be slightly better for domain adaptation. Possibly, this is due to increased modelling capacity. This benefit wanes as the discrepancy between the training/test sample increases. As before, the cause may be overfitting since overfitting and class imbalance are known problems in discourse parsing \cite{atwell2021we}.

\paragraph{News Test Set}
\emph{It is slightly harder to transfer to news datasets}. We consider a ``news'' corpus to be any of PDTB, RST-DT, or the news domain of GUM. When the target (test) dataset consists of news texts, we see adaptation performance consistent with non-news targets for small discrepancy. As the discrepancy between training and test set grows, the non-news targets are actually better suited for domain adaptation; i.e., it is slightly easier to transfer to a non-news target. Possibly, this is related to the length and complexity of news texts. 

\paragraph{Dataset}
\emph{Increased variability in the target domain results in a more difficult task, even when adding variability during training.} In general, we see that the GUM dataset presents a more challenging adaptation task than the other datasets. This is sensible due to the larger selection of target domains in GUM. 
In our results, increased variability at train-time does not appear to counteract this issue, because adaptation experiments in the GUM corpus are multi-source.
For PDTB, as the discrepancy increases, performance is more similar to GUM. On the other hand, RST-DT presents the easiest adaptation task. This is expected as all test sets in the RST-DT experiments 
are drawn from the same news corpus.

%% file: 06-conclusion.tex
\section{Conclusion}
\label{sec:conclusion}
This work provides a statistic for model and dataset selection, that we also use to conduct large-scale analysis of model transfer in discourse parsing. Our analysis provides useful insights for the practitioner.
For one, 
the correlations indicate
that, for datasets with the RST-DT annotation framework, the statistics that quantify distributional shift without being directly tied to error-gap (where error-gap refers to the performance gap between train and test splits) are very weakly correlated with error-gap. This also holds for non-news targets, and indicates that the $h$-discrepancy is especially useful for predicting the effects of domain shift in these cases.

Additionally, 
we find that:
(1) increased variability in the target domain appears to make domain adaptation more difficult, even if the training set contains a similar level of variability; 
(2) S-BERT is better than A-BERT 
when domains are similar, but A-BERT outperforms S-BERT when the domains further diverge; (3) non-news texts (such as those in the BioDRB) are easier to adapt to than news texts (such as those in the PDTB).

This is the first computational and empirical study that looks at distribution shifts across different
discourse datasets and evaluates the performance of various models under these shifts. This is also the first work that examines the efficacy of different two-sample tests for predicting the error-gap when compared to a metric that is theoretically tied to error gap. Future work can extend these results by using the \emph{h}-discrepancy metric to predict the error-gap for other NLP tasks 
or for other components needed for discourse parsing, such as constructing the RST-DT dataset. 

%% file: 07-appendix_a.tex
\section{Frameworks}
\label{sec:frameworks}

The Penn Discourse Treebank \cite{PDTB2004, PDTB2008, webber2019penn} consists of Wall Street Journal articles labeled with both \emph{explicit} and \emph{implicit} shallow discourse relations (relations between only two text units). 
Explicit discourse relations are ones in which a \textit{connective} between the arguments provides some indication of the correct discourse sense label. Implicit discourse relations, which we focus on in this paper, are ones in which a connective can be \textit{inserted} that indicates the correct sense.

The RST Discourse Treebank \cite{carlson2003building} is a corpus containing Wall Street Journal articles annotated in the style of Rhetorical Structure Theory, where a document is split into \emph{elementary discourse units} (EDUs) and relations made up of these EDUs form a tree structure.  The RST Discourse Treebank does not differentiate between explicit and non-explicit discourse relations, nor does it label discourse connectives. 

%% file: 10-appendix_d.tex
\section{Model Training and Transfer Results}
\label{sec:model-training}

\paragraph{Optimization Parameters} We use SGD on an NLL loss with momentum set to $0.9$ to train all of our models. We use a batch size of $250$. We start training with a learning of $1 \times 10^{-2}$ for 100 epochs and then train for another 50 epochs using a learning rate of $1 \times 10^{-3}$. If a model achieves a training error lower than $5 \times 10^{-4}$, we stop training. 

\begin{figure*}
    \centering
    \includegraphics[width=\textwidth]{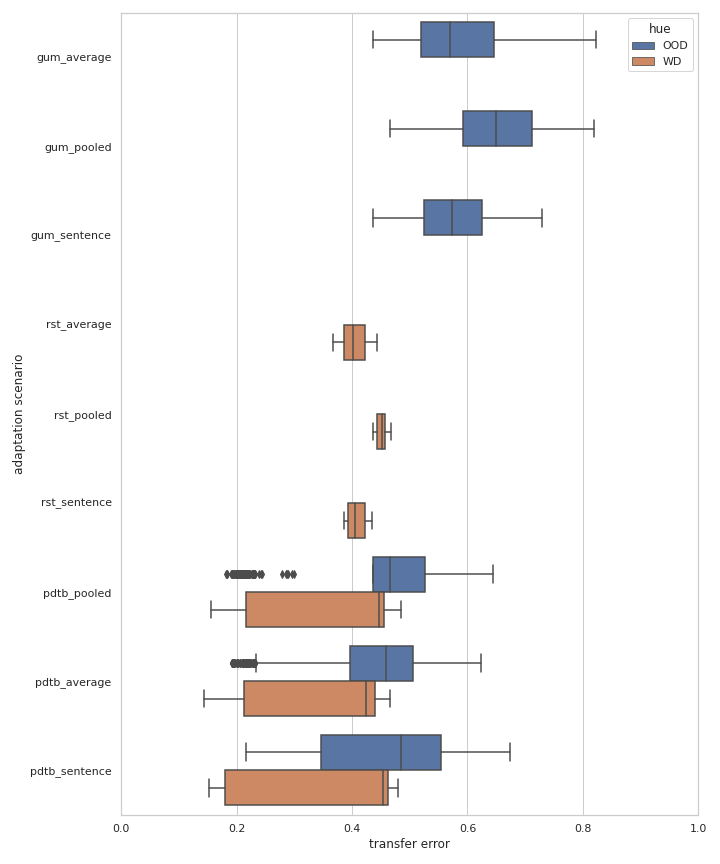}
    \caption{Transfer error within and out of distribution for each dataset}
    \label{fig:transfer_error_wd_ood}
\end{figure*}

\begin{figure*}
    \centering
    \includegraphics[width=\textwidth]{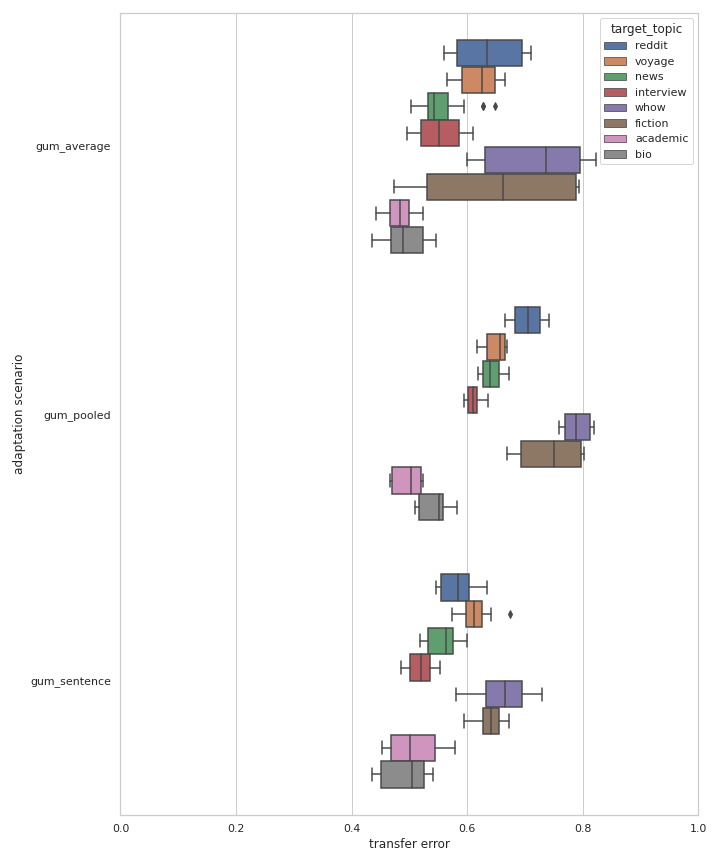}
    \caption{Transfer error for each topic within the GUM corpus}
    \label{fig:transfer_error_gum}
\end{figure*}

%% file: 11-appendix_e.tex
\section{Two-Sample Statistics}
\label{sec:two-sample-details}
Here, we describe in detail the common two-sample statistics listed in Section~\ref{sec:methods} and studied in Section~\ref{sec:results}
\paragraph{Friedman-Rafsky Test Statistic}
The Friedman-Rafsky Test Statistic $R$ \cite{friedman1979multivariate} is computed by forming a minimum-spanning tree (MST) using the pooled sample $P = (X_i \mid (X_i, Y_i) \in S) +_\mathrm{c} (\tilde{X}_i \mid \tilde{X}_i \in T_X)$ of marginal features. Here, $+_\mathrm{c}$ is the concatenation operation. To form the tree, we form a weighted graph $G_P$ by treating each point $Z_i \in P$ as vertex and assigning an edge between each pair of vertices whose weight is the distance between the data-points. When $\mathcal{X} = \mathbb{R}^d$ for some $d$, this is usually the Euclidean distance or L2 norm. The MST is then precisely the MST of $G_P$. The statistic $R$ is computed as the number of edges whose endpoints originally belonged to the same sample. For example, $R$ increases by 1 for each edge whose endpoints both originally belong to $T_X$. Likewise, $R$ increases by 1 for each edge whose endpoints are both the features of points in $S$. When endpoints originally belonged to distinct samples, $R$ remains unmodified. We report modified statistic below which is normalized to account for sample size $R_\mathrm{normed} = R / (n + m - 2)$. Since the size of the MST is $n + m - 1$ and there is always at least one edge between $S$ and $T_X$, this statistic has a maximum value of $1$.


\paragraph{Energy Statistic}
Given samples $S$ and $T_X$ as before, the energy statistic may be computed as below
\begin{equation}\small
\begin{split}
E = & \frac{2}{nm} \sum\nolimits_{i,j} ||X_i - \tilde{X}_j|| - \frac{1}{n^2} \sum\nolimits_{i,j} ||X_i - X_j||  \\
& - \frac{1}{m^2} \sum\nolimits_{i,j} ||\tilde{X}_i - \tilde{X}_j||
\end{split}
\end{equation}
where $||\cdot||$ gives the Euclidean norm (distance). Originally proposed by \citet{szekely2013energy}, the statistic is motivated by Newton's potential energy between heavenly bodies. Intuitively, it is fairly easy to understand as a comparison of dissimilarity within samples and across samples. If the dissimilarity across samples (i.e., the first term) is much higher than the dissimilarity within samples, then the two samples are likely drawn from different distributions.  

\paragraph{Maximum Mean Discrepancy (MMD)} 
Given samples $S$ and $T_X$ as before, the MMD statistic \cite{gretton2012kernel} may be computed as below
\begin{equation}\small
\begin{split}
M = & \frac{\sum\nolimits_{i\neq j} K(X_i, X_j)}{n(n-1)} 
+ \frac{\sum\nolimits_{i\neq j} K(\tilde{X}_i, \tilde{X}_j)}{m(m-1)}  \\
& - \frac{2}{nm} \sum\nolimits_{i,j} K(X_i, \tilde{X}_j)
\end{split}
\end{equation}
where $K : \mathcal{X} \times \mathcal{X} \to \mathbb{R}_{\geq 0}$ is the kernel for some RKHS. In our experiments, we use an Gaussian RBF kernel and select $\sigma$ to be an approximate\footnote{Specifically, we use a smaller random sample of 100 data points to compute this median.} median distance of the pooled sample as done by \citet{rabanser2019failing}. Intuitively, $K$ behaves as a similarity metric between points in $\mathcal{X}$ and, in this sense, the MMD statistic compares samples in much the same way that the energy statistic does. Rather than dissimilarity, the MMD statistic looks at similarity of points within and across samples, modifying the order of the summands appropriately to retain direct proportionality with the difference in samples.

%% file: 12-appendix_f.tex
\section{Proof of Theorem~\ref{thm:main}}
\label{sec:proof}
\begin{proof}
We use the triangle inequality of classification error \cite{crammer2007learning, ben2007analysis}. For any realization of the sample $S$ and any distribution $\mathbb{T}$ over $\mathcal{X} \times \mathcal{Y}$, for any classifiers $h, h' \in \mathcal{H}$, the triangle inequality yields\footnote{A full derivation of Eq.~\eqref{eqn:ben-david} may be found by following steps as in the proof of Theorem~2 of \citet{ben2010theory}:
\begin{equation}
    \begin{split}
        & \mathbf{R}_\mathbb{T}(h)
\leq \mathbf{R}_\mathbb{T}(h, h') + \mathbf{R}_\mathbb{T}(h') \\
& \leq \mathbf{R}_S(h,h') + \mathbf{R}_\mathbb{T}(h') + |\mathbf{R}_\mathbb{T}(h, h') - \mathbf{R}_S(h, h')| \\
& \leq \mathbf{R}_S(h) + \mathbf{R}_S(h') + \mathbf{R}_\mathbb{T}(h') + |\mathbf{R}_\mathbb{T}(h, h') - \mathbf{R}_S(h, h')|
    \end{split}
\end{equation}}
\begin{equation}\label{eqn:ben-david}
\begin{split}
    \mathbf{R}_\mathbb{T}(h) - & \mathbf{R}_S(h) \leq \mathbf{R}_S(h') + \mathbf{R}_\mathbb{T}(h') \\
    & + \lvert \mathbf{R}_S(h,h') - \mathbf{R}_\mathbb{T}(h,h') \rvert
\end{split}
\end{equation}
where for $\mathbb{T}$ over $\mathcal{X} \times \mathcal{Y}$ we have
\begin{equation}
    \mathbf{R}_\mathbb{T}(h,h') = \underset{\tilde{X} \sim \mathbb{T}_X}{\mathbf{Pr}}(h(\tilde{X}) \neq h'(\tilde{X}))
\end{equation}
and for $S = (X_i, Y_i)_{i=1}^n$ we have
\begin{equation}
    \mathbf{R}_S(h,h') = n^{-1}\sum_{i=1}^n 1[h(X_i) \neq h'(X_i)].
\end{equation}

Interchanging roles of $\mathbb{T}$ and $S$ in Eq.~\eqref{eqn:ben-david} and using the definition of the absolute value, we see
\begin{equation}
\begin{split}
    \Delta_h(S, \mathbb{T}) \leq \ & \mathbf{R}_S(h') + \mathbf{R}_\mathbb{T}(h') \\
    & + \lvert \mathbf{R}_S(h,h') - \mathbf{R}_\mathbb{T}(h,h') \rvert.
\end{split}
\end{equation}
For brevity, for any distribution $\mathbb{D}$, set 
\begin{equation}
    \xi(\mathbb{D}) = \lvert \mathbf{R}_S(h,h') - \mathbf{R}_\mathbb{D}(h,h') \rvert.
\end{equation}
Then, using the common ``addition of zero'' trick, we arrive at 
\begin{equation}
\begin{split}
    \Delta_h(S, \mathbb{T}) & \leq \mathbf{R}_S(h') + \mathbf{R}_\mathbb{T}(h') \\
    & - \mathbf{R}_T(h') + \mathbf{R}_T(h') + \xi(\mathbb{T}) \\
    & - \xi(T) + \xi(T).
\end{split}
\end{equation}
Then, by monotonicity and linearity of the expectation we have
\begin{equation}\label{eqn:2}
\begin{split}
    \Delta_h(S, \mathbb{T}) & \leq \mathbf{E}_T \Big [ \mathbf{R}_S(h') + \mathbf{R}_T(h') \Big ] \\
    & + \mathbf{E}_T \big [ \xi(T)\big ] \\
    & + \mathbf{R}_\mathbb{T}(h') - \mathbf{E}_T \Big [ \mathbf{R}_T(h') \Big ] \\
    & + \xi(\mathbb{T}) - \mathbf{E}_T \big [ \xi(T)\big].
\end{split}
\end{equation}
Let us consider some of these terms individually. Using linearity of expectation and the correspondence between probability and the expectation of an indicator function, we have 
\begin{equation}\label{eqn:unbiased}
    \begin{split}
        & \mathbf{E}_T \Big [ \mathbf{R}_T(h') \Big ] = \mathbf{E} \Bigg [ m^{-1} \sum_{i=1}^m 1[h(\tilde{X}_i) \neq \tilde{Y}_i]\Bigg ] \\
        & =  m^{-1} \sum_{i=1}^m \mathbf{E} \big [ 1[h(\tilde{X}_i) \neq \tilde{Y}_i] \big ] \\
        & = m^{-1} \sum_{i=1}^m  \underset{(\tilde{X}_i, \tilde{Y}_i) \sim \mathbb{T} }{\mathbf{Pr}} \big (h(\tilde{X}_i) \neq \tilde{Y}_i\big ) \\
        & = m^{-1} \sum_{i=1}^m \mathbf{R}_\mathbb{T}(h) \\
        & = \mathbf{R}_\mathbb{T}(h).
    \end{split}
\end{equation}
Additionally, we have
\begin{equation}
\begin{split}
& \mathbf{E}_T \big [ \xi(T)\big] = \mathbf{E}_T\Big [\lvert \mathbf{R}_S(h,h') - \mathbf{R}_T(h,h') \rvert \Big] \\
& \geq \lvert \mathbf{R}_S(h,h') - \mathbf{E}\big [\mathbf{R}_T(h,h') \big] \rvert \\
& = \xi(\mathbb{T}).
\end{split}
\end{equation}
Here, the second line follows by Jensen's Inquality and linearity of the expectation. The last line follows using a similar derivation as in Eq.~\eqref{eqn:unbiased}. Then, 
\begin{equation}
    \xi(\mathbb{T}) - \mathbf{E}_T[\xi(T)] \leq 0
\end{equation}
and 
\begin{equation}
    \mathbf{R}_\mathbb{T}(h') - \mathbf{E}_T \Big [ \mathbf{R}_T(h') \Big ] = 0.
\end{equation}
Using these two facts in conjunction with Eq.~\eqref{eqn:2} yields
\begin{equation}
\begin{split}
    \Delta_h(S, \mathbb{T}) & \leq \mathbf{E}_T \Big [ \mathbf{R}_S(h') + \mathbf{R}_T(h') \Big ] \\
    & + \mathbf{E}_T \big [ \xi(T)\big ].
\end{split}
\end{equation}
Using $h$ as in Eq.~\eqref{eqn:D} to define the statistic $D$, for any $h' \in \mathcal{H}$, we know $\xi(T) \leq D$ (i.e., by definition of $\mathrm{max}$). So, monotonicity and linearity of expectation implies $\mathbf{E}_T [\xi(T)] \leq \mathbf{E}_T[D]$. For an appropriate choice of $h'$, we then have
\begin{equation}
\begin{split}
    \Delta_h(S, \mathbb{T}) & \leq \mathbf{E}_T \big [ \lambda \big ] + \mathbf{E}_T \big [ D \big ].
\end{split}
\end{equation}
Rearranging terms gives the lowerbound and the upperbound follows immediately from the fact that $\Delta_h(S, \mathbb{T})$ is non-negative. 

\end{proof}

%% file: 13-appendix_g.tex
\section{Regression Diagnostics}
\label{sec:reg-diag}
\begin{figure}[t]
    \centering
    \includegraphics[width=\columnwidth]{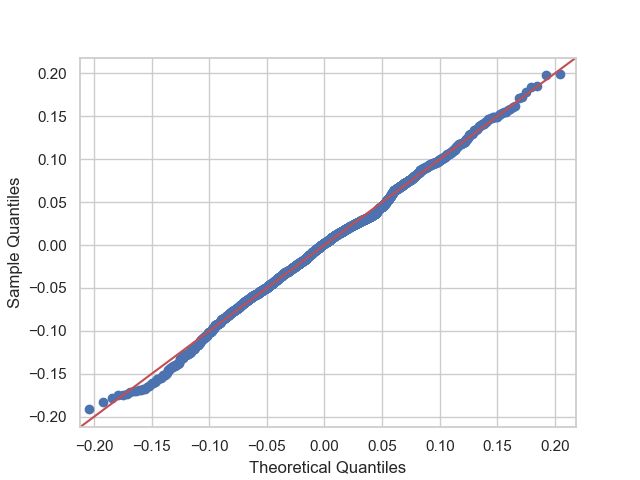}
    \caption{Quantile-Quantile plot. Red line shows ideal: sample quantiles should be the same as the theoretical quantiles of a normal distribution with same variance.}
    \label{fig:qqplot}
\end{figure}
\begin{figure}[t]
    \centering
    \includegraphics[width=\columnwidth]{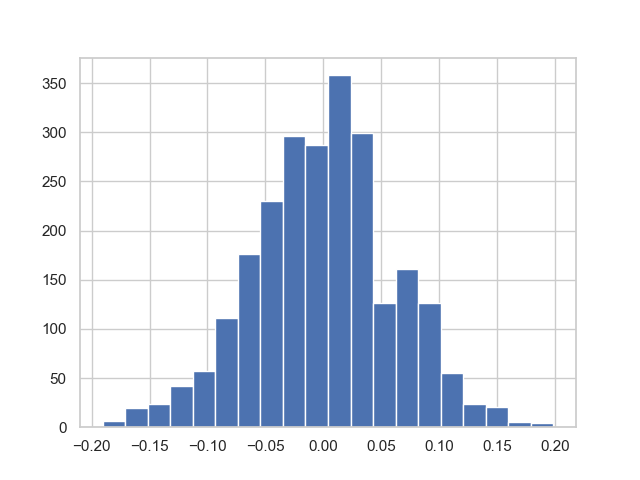}
    \caption{ Histogram of realized error terms. Horizontal axis shows value of error term, while vertical axis shows count.}
    \label{fig:resids_hist}
\end{figure}
\begin{figure*}
    \centering
    \includegraphics[width=\textwidth]{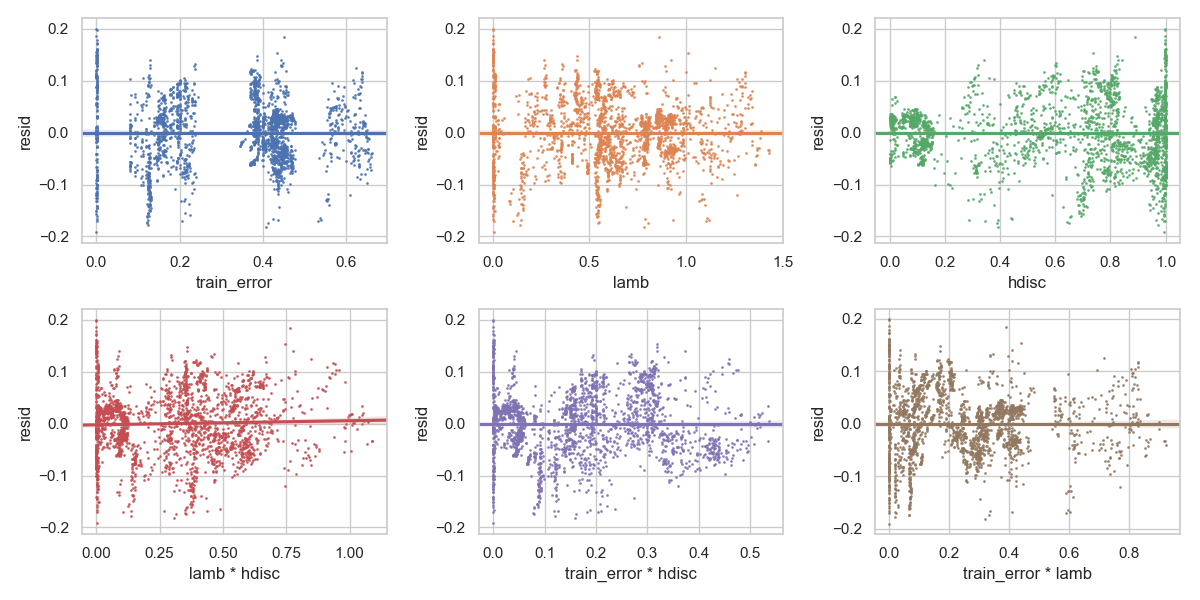}
    \caption{ Residual plots. Vertical axes show realized error terms, while horizontal axes show value of some feature that may or may not be in our design matrix. Significant patterns may indicate a missing term in our model. While some patterning may exist, we choose not to include additional terms for reason of interpretability and to meet other (quantifiable) model assumptions.}
    \label{fig:resids}
\end{figure*}

\paragraph{Normal Errors Assumption} Here, we give diagnostics for the regression model used to analyze data in the main text. Primarily, we would like to check the assumptions that our error terms (i.e., $\boldsymbol\epsilon$) are all identically and independently normally distributed. The Jarque-Bera (JB) test uses a statistic based on the skew and kurtosis of the observed errors to study this hypothesis. Assuming the residuals are i.i.d. normal, the probability of observing a JB statistic as extreme as observed is $\approx 0.25$. So, we fail to reject the hypothesis that the residuals are i.i.d normal at significance level $\alpha = 0.05$. The assumption that error terms are normal distributed may also be visually checked using the qq-plot, histogram of errors, and the residual plots contained in Figures~\ref{fig:qqplot}, \ref{fig:resids_hist}, and \ref{fig:resids}, respectively. We do not see particularly strong evidence that the residuals are not i.i.d. normal. Albeit, some patterning in the residual plots and skew in the histogram of residuals may be of concern.

\paragraph{Other Possible Assumptions} In any case, even if the normality assumption does not hold, our analysis can still be interpreted using more loose assumptions. The most important assumption is that the error terms all have mean 0. Empirically, we find this to be the case with the average residual being $\approx 2.4 \times 10^{-15}$. In fact, Figure~\ref{fig:resids} shows the line-of-best fit through the residuals (which is typically close to the zero line). As long as the assumption that the error terms have common mean 0 is true, the OLS estimates we use for the coefficients will be unbiased. The only possible short-coming of the OLS estimate is that it could have larger variance than some other estimate. In our analysis, we are most concerned with the unbiased property of our coefficient estimates, but a larger variance in our estimator decreases our confidence that this particular experiment produces estimates close to the truth. Either way, under our relaxed assumption of only a common mean 0 in the errors, we can expect our analysis in the main text to reveal the truth across repeated experiments.

\begin{table*}
\begin{center}
\begin{tabular}{lclc}
\toprule
\textbf{Dep. Variable:}                  &     est. error   & \textbf{  R-squared:         } &     0.944   \\
\textbf{Model:}                          &       OLS        & \textbf{  Adj. R-squared:    } &     0.944   \\
\textbf{Method:}                         &  Least Squares   & \textbf{  F-statistic:       } &     1949.   \\
                                         &                  & \textbf{  Prob (F-statistic):} &     0.00    \\
                                         &                  & \textbf{  Log-Likelihood:    } &    3347.1   \\
\textbf{No. Observations:}               &        2428      & \textbf{  AIC:               } &    -6650.   \\
\textbf{Df Residuals:}                   &        2406      & \textbf{  BIC:               } &    -6523.   \\
\textbf{Df Model:}                       &          21      & \textbf{                     } &             \\
\bottomrule
\end{tabular}
\begin{tabular}{lcccccc}
                                         & \textbf{coef} & \textbf{std err} & \textbf{t} & \textbf{P$> |$t$|$} & \textbf{[0.025} & \textbf{0.975]}  \\
\midrule
\textbf{Intercept}                       &      -0.0206  &        0.034     &    -0.606  &         0.545        &       -0.087    &        0.046     \\
\textbf{hspace[T.lin]}                   &      -0.0239  &        0.006     &    -3.817  &         0.000        &       -0.036    &       -0.012     \\
\textbf{group[T.pdtb]}                   &       0.0536  &        0.016     &     3.340  &         0.001        &        0.022    &        0.085     \\
\textbf{group[T.rst]}                    &       0.0600  &        0.018     &     3.256  &         0.001        &        0.024    &        0.096     \\
\textbf{bert[T.pooled]}                  &       0.0034  &        0.006     &     0.601  &         0.548        &       -0.008    &        0.015     \\
\textbf{bert[T.sentence]}                &       0.0250  &        0.009     &     2.872  &         0.004        &        0.008    &        0.042     \\
\textbf{news[T.notnews]}                 &      -0.0029  &        0.010     &    -0.289  &         0.773        &       -0.022    &        0.017     \\
\textbf{train\_error}                    &       0.3262  &        0.080     &     4.054  &         0.000        &        0.168    &        0.484     \\
\textbf{lamb}                            &      -0.0150  &        0.048     &    -0.312  &         0.755        &       -0.109    &        0.079     \\
\textbf{hdisc}                           &       0.1545  &        0.081     &     1.906  &         0.057        &       -0.004    &        0.313     \\
\textbf{bert[T.pooled]:hdisc}            &      -0.0313  &        0.009     &    -3.622  &         0.000        &       -0.048    &       -0.014     \\
\textbf{bert[T.sentence]:hdisc}          &      -0.1370  &        0.013     &   -10.600  &         0.000        &       -0.162    &       -0.112     \\
\textbf{hspace[T.lin]:hdisc}             &       0.0194  &        0.009     &     2.159  &         0.031        &        0.002    &        0.037     \\
\textbf{group[T.pdtb]:hdisc}             &      -0.0210  &        0.021     &    -1.002  &         0.316        &       -0.062    &        0.020     \\
\textbf{group[T.rst]:hdisc}              &       0.0671  &        0.028     &     2.410  &         0.016        &        0.013    &        0.122     \\
\textbf{news[T.notnews]:hdisc}           &       0.0320  &        0.013     &     2.529  &         0.012        &        0.007    &        0.057     \\
\textbf{hdisc:train\_error}              &       1.9665  &        0.196     &    10.052  &         0.000        &        1.583    &        2.350     \\
\textbf{np.power(hdisc, 2)}              &       0.4831  &        0.052     &     9.323  &         0.000        &        0.381    &        0.585     \\
\textbf{train\_error:np.power(hdisc, 2)} &      -1.6867  &        0.152     &   -11.074  &         0.000        &       -1.985    &       -1.388     \\
\textbf{lamb:train\_error}               &      -0.5861  &        0.122     &    -4.803  &         0.000        &       -0.825    &       -0.347     \\
\textbf{np.power(lamb, 2)}               &      -0.1346  &        0.071     &    -1.892  &         0.059        &       -0.274    &        0.005     \\
\textbf{train\_error:np.power(lamb, 2)}  &       0.4043  &        0.100     &     4.029  &         0.000        &        0.208    &        0.601     \\
\bottomrule
\end{tabular}
\begin{tabular}{lclc}
\textbf{Omnibus:}       &  2.707 & \textbf{  Durbin-Watson:     } &    1.548  \\
\textbf{Prob(Omnibus):} &  0.258 & \textbf{  Jarque-Bera (JB):  } &    2.718  \\
\textbf{Skew:}          & -0.046 & \textbf{  Prob(JB):          } &    0.257  \\
\textbf{Kurtosis:}      &  3.136 & \textbf{  Cond. No.          } &     463.  \\
\bottomrule
\end{tabular}
\end{center}
Warnings: \newline
 [1] Standard Errors assume that the covariance matrix of the errors is correctly specified.
    \caption{ Full description of the regression model including all features, estimated coefficients, and relevant tests for diagnosis and inference. Tests involving standard errors (std err) are only valid if the model errors follow the assumed distribution. We believe most variables are self-explanatory, but we do provide some assistance to reader: \textbf{lamb} corresponds to $\lambda$, \textbf{hdisc} corresponds to the $h$-discrepancy, \textbf{train\_error} corresponds to the error on the source sample, \textbf{np.power($\diamond$, 2)} corresponds to the square of the feature $\diamond$, presence of \textbf{:} indicates a multiplication of features (i.e., an interaction-term), and \textbf{hspace} corresponds to the type of classifier used (i.e., linear model or fully-connected network).}
    \label{tab:reg-diag}
\end{table*}

%% file: 14-appendix_h.tex
\section{Regression Analysis Examples}
\label{sec:reg-eg}
In this section, we give detailed examples (i.e., Exampled~\ref{examp:reg-1} and \ref{examp:reg-2}) to clarify how we compute estimates in Figure~\ref{fig:all}. As noted, we use the unbiased OLS estimate $\boldsymbol{\hat{\beta}} = (\mathbf{X}^\mathrm{T}\mathbf{X})^{-1}\mathbf{X}^\mathrm{T}\mathbf{Y}$ in place of $\boldsymbol\beta$ as is standard.
\begin{example}
\label{examp:reg-1}
Let column $j$ of $\mathbf{X}$ contain the realizations of the $h$-discrepancy for each experiment and let column $k$ contain the train error. Suppose column $\ell$ is the (element-wise) product of columns $k$ and $j$, column $q$ is the square of column $j$, and column $r$ is the product of columns $q$ and $k$. Then, controlling for all other features in $\mathbf{X}$, the expected change in estimation error per $\delta > 0$ increase in the $h$-discrepancy is 
\begin{equation}\small
\begin{split}
    & \mathbf{E}[\mathbf{Y}_i \mid \mathbf{X}_i = \mathbf{x}] - \mathbf{E}[\mathbf{Y}_i \mid \mathbf{X}_i = \mathbf{x}']
    = \beta_j \delta + \beta_\ell \delta \mathbf{x}'_k \\
    & + \beta_q (\delta^2 + 2\delta \mathbf{x}'_j) + \beta_r (\delta^2\mathbf{x}'_k + 2\delta \mathbf{x}'_j\mathbf{x}'_k)
\end{split}
\end{equation}
where $\mathbf{x}'$ is a fixed row-vector of features and $\mathbf{x}$ is defined by
\begin{equation}
    \mathbf{x}_p = \begin{cases}
    \mathbf{x}'_p + \delta & \text{if} \ p = j, \\
    \mathbf{x}'_k (\mathbf{x}'_j + \delta) & \text{if} \ p = \ell, \\
    (\mathbf{x}'_j + \delta)^2 & \text{if} \ p = q, \\
    \mathbf{x}'_k (\mathbf{x}'_j + \delta)^2 & \text{if} \ p = r, \\
    \mathbf{x}'_p & \text{else}
    \end{cases}.
\end{equation}
If this function of $\delta$ is positive, we know increasing the $h$-discrepancy increases the bias as suggested by our theory.
\end{example}

\begin{example}\label{examp:reg-2}
Let column $j$ of $\mathbf{X}$ be 1 if we use S-BERT representations and 0 otherwise. Let column $k$ of $\mathbf{X}$ indicate use of P-BERT in the same way and suppose the reference category\footnote{In regression, the reference is the single category from any group of categories which is not explicitly included in $\mathbf{X}$. It serves as a point of comparison for the other categories. For technical reasons, a point of comparison is typically needed to analyze impact of categorical features (i.e., so $\mathbf{X}$ is full rank).} for the BERT representations is A-BERT. Let column $\ell$ of $\mathbf{X}$ contain discrepancy $D_i$ for each experiment and let column $q$ be the element-wise product of columns $j$ and $\ell$; i.e., interaction terms. Then, controlling for all other features in $\mathbf{X}$, the expected increase in error-gap using S-BERT instead of A-BERT is 
\begin{equation}\small
\begin{split}
    & \mathbf{E}[D_i - \mathbf{Y}_i \mid \mathbf{X}_i = \mathbf{x}] - \mathbf{E}[D_i - \mathbf{Y}_i \mid \mathbf{X}_i = \mathbf{x}'] \\
    & = -(\beta_j + \beta_q D_i)
\end{split}
\end{equation}
where $\mathbf{x}'$ is a fixed row-vector of features such that $\mathbf{x}'_\ell = D_i$ and $\mathbf{x}'_j = \mathbf{x}'_k = 0$. The row-vector $\mathbf{x}$ is defined by $\mathbf{x}_r = \{1 \ \mathrm{if} \ r = j, \ \mathbf{x}'_\ell \ \mathrm{if} \ r = q, \ \mathbf{x}'_r \ \mathrm{else}\}$. When this function of $D_i$ is positive, we know using S-BERT is expected to increase the error-gap.
\end{example}